%% file: main.tex
\documentclass[10pt,twocolumn,letterpaper]{article}

\usepackage{iccv}
\usepackage{times}
\usepackage{epsfig}
\usepackage{graphicx}
\usepackage{amsmath}
\usepackage{amsthm}
\usepackage{amssymb}
\usepackage{booktabs}
\usepackage{tablefootnote}
\usepackage{pifont}
\usepackage[linesnumbered,ruled,vlined]{algorithm2e}
\usepackage{bm}

\usepackage{xcolor}

\DeclareMathOperator*{\argmax}{arg\,max}

\newcommand\mat[1]{\mathbf{#1}}
\newcommand\mati[2]{\mat{#1}^{(#2)}}
\renewcommand\vec[1]{\mathbf{#1}}
\newcommand\veci[2]{\vec{#1}^{(#2)}}

\newcommand\W{\mat{W}}
\newcommand\Wt{\mat{\widetilde{W}}}
\newcommand\xx{\vec{x}}
\newcommand\y{\vec{y}}

\newcommand\yt{\vec{\widetilde{y}}}
\newcommand\beps{\boldsymbol{\epsilon}}
\newcommand{\bgam}{\boldsymbol{\gamma}}
\newcommand{\bbet}{\boldsymbol{\beta}}
\newcommand{\mutrunc}{{\mu^t_{ab}}}

\newcommand{\muclip}{{\mu^c_{ab}}}
\newcommand{\sigclip}{{\sigma^c_{ab}}}

\newcommand{\good}{{\color{green}\checkmark}}
\newcommand{\bad}{{\color{red}\ding{55}}}

\usepackage[breaklinks=true,bookmarks=false]{hyperref}

\iccvfinalcopy 

\def\iccvPaperID{6174} 

\ificcvfinal\fi
\begin{document}


\title{Data-Free Quantization\\ Through Weight Equalization and Bias Correction}

\author{Markus Nagel\thanks{Equal Contribution}\\
\and
Mart van Baalen\footnotemark[1]\\
\and
Tijmen Blankevoort\\
\and
Max Welling\\
\and
Qualcomm AI Research\thanks{Qualcomm AI Research is an initiative of Qualcomm Technologies, Inc.}\\
Qualcomm Technologies Netherlands B.V.\\
{\tt\small \{markusn, mart, tijmen, mwelling\}@qti.qualcomm.com}\\
}


\maketitle
\ificcvfinal\thispagestyle{empty}\fi

\begin{abstract}
\input{sections/abstract.tex}
\end{abstract}

\vspace{-0.3cm}
\section{Introduction}\label{sec:introduction}
\input{sections/introduction.tex}

\section{Background and related work}\label{sec:backgroundrelated}
\input{sections/related_work.tex}

\section{Motivation}\label{sec:motivation}
\input{sections/motivation.tex}

\section{Method}\label{sec:method}
\input{sections/method.tex}

\section{Experiments}\label{sec:experiments}
\input{sections/experiments.tex}

\section{Conclusion}\label{sec:conclusion}
\input{sections/conclusion.tex}

\section*{Acknowledgments}\label{sec:acknowledgments}
We would like to thank Christos Louizos, Harris Teague, Jakub Tomczak, Mihir Jain and Pim de Haan for their helpful discussions and valuable feedback.

{\small
\bibliographystyle{ieee_fullname}
\bibliography{references}
}

\newpage
\clearpage

\appendix
\section{Optimal range equalization of two layers}\label{app:equalization}
\input{sections/appendix-cle.tex}

\section{Bias correction for convolutional layers}\label{app:biascorr}
\input{sections/appendix-biascorr.tex}

\section{Clipped normal distribution}\label{app:clippednormal}
\input{sections/appendix-clipnorm.tex}
\section{Empirical quantization bias correction}
\input{sections/appendix-empirical-quant-bias-correction.tex}\label{app:databias}

\section{Additional experiments}
\input{sections/appendix-additional-results.tex}\label{app:addres}

\end{document}

%% file: sections/abstract.tex
We introduce a data-free quantization method for deep neural networks that does not require fine-tuning or hyperparameter selection. 
It achieves near-original model performance on common computer vision architectures and tasks. 
8-bit fixed-point quantization is essential for efficient inference on modern deep learning hardware. 
However, quantizing models to run in 8-bit is a non-trivial task, frequently leading to either significant performance reduction or engineering time spent on training a network to be amenable to quantization. 
Our approach relies on equalizing the weight ranges in the network by making use of a scale-equivariance property of activation functions. 
In addition the method corrects biases in the error that are introduced during quantization. 
This improves quantization accuracy performance, and can be applied to many common computer vision architectures with a straight forward API call.
For common architectures, such as the MobileNet family, we achieve state-of-the-art quantized model performance. 
We further show that the method also extends to other computer vision architectures and tasks such as semantic segmentation and object detection.

%% file: sections/introduction.tex
In recent years, deep learning based computer vision models have moved from research labs into the cloud and onto edge devices.
As a result, power consumption and latency of deep learning inference have become an important concern.
For this reason fixed-point quantization is often employed to make inference more efficient.
By quantizing floating point values onto a regularly spaced grid, the original floating point values can be approximated by a set of integers, a scaling factor, and an optional zero point offset \cite{jacob2018cvpr}.
This allows for the use of faster and more power-efficient integer operations in matrix multiplication and convolution computations, at the expense of lower representational power.
We refer the reader to \cite{krishnamoorthi} for details on commonly used, hardware-friendly quantization methods for deep learning models.

\begin{figure}[t]
\includegraphics[width=8cm, trim={.2cm .2cm .2cm .7cm}, clip]{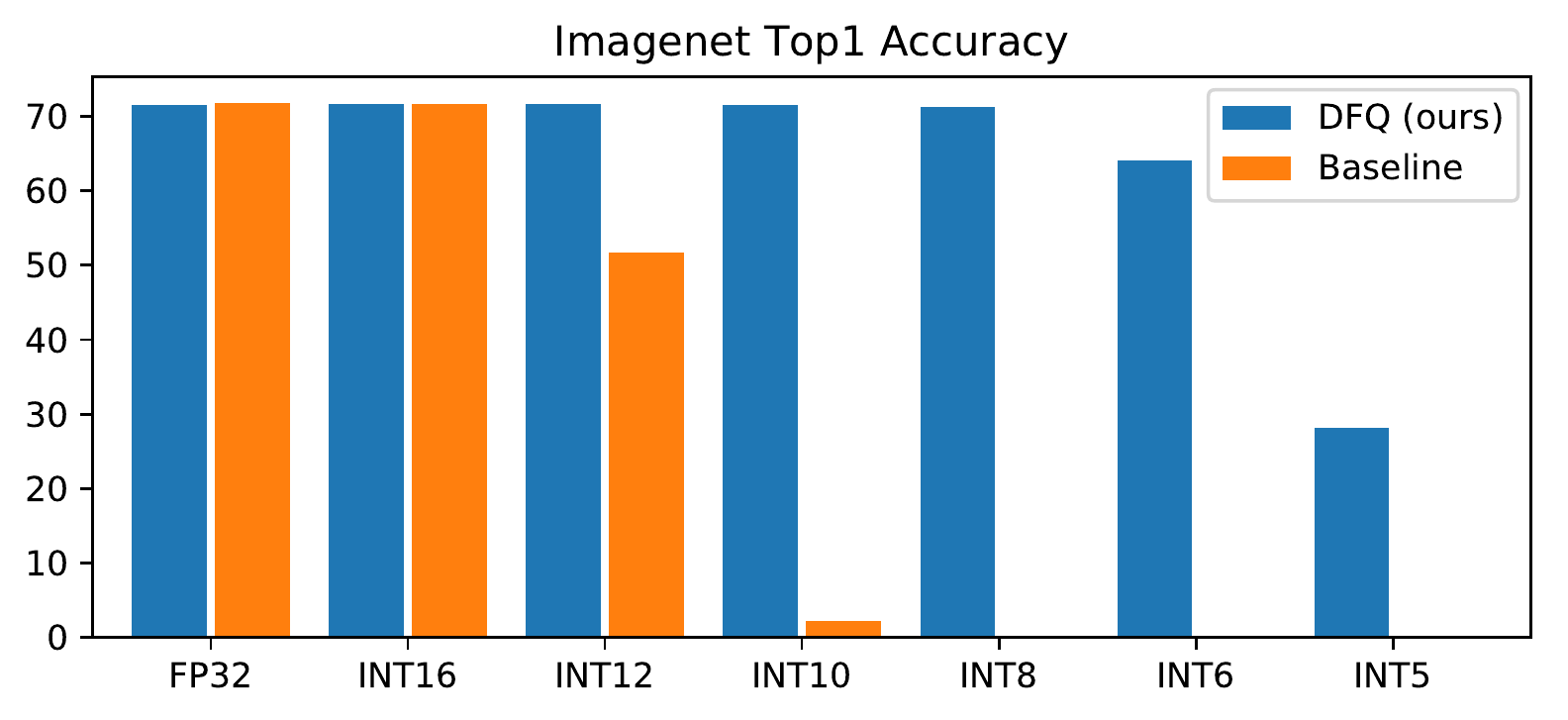}
\centering
\caption{Fixed point inference for MobileNetV2 on ImageNet. The original model has significant drop in performance at 12-bit quantization whereas our model maintains close to FP32 performance even at 6-bit quantization.}
\label{fig:imagenet_comparison}
\end{figure}

Quantization of 32-bit full precision (FP32) models into 8-bit fixed point (INT8) introduces quantization noise on the weights and activations, which often leads to reduced model performance.
This performance degradation ranges from very minor to catastrophic.
To minimize the quantization noise, a wide range of different methods have been introduced in the literature (see Section \ref{sec:backgroundrelated}).
A major drawback of these quantization methods is their reliance on data and fine-tuning. 
As an example, consider real-world actors that manage hardware for quantized models, such as cloud-based deep learning inference providers or cellphone manufacturers.
To provide a general use quantization service they would have to receive data from the customers to fine-tune the models, or rely on their customers to do the quantization. 
In either case, this can add a difficult step to the process. 
For such stakeholders it would be preferable if FP32 models could be converted directly to INT8, without needing the know-how, data or compute necessary for running traditional quantization methods. 
Even for model developers that have the capability to quantize their own models, automation would save significant time.
%

In this paper, we introduce a quantization approach that does not require data, fine-tuning or hyperparameter tuning, resulting in accuracy improvement with a simple API call. 
Despite these restrictions we achieve near-original model performance when quantizing FP32 models to INT8. 
This is achieved by adapting the weight tensors of pre-trained models such that they are more amenable to quantization, and by correcting for the bias of the error that is introduced when quantizing models.
We show significant improvements in quantization performance on a wide range of computer vision models previously thought to be difficult to quantize without fine-tuning. 

\paragraph{Levels of quantization solutions}\vspace{-.1cm}
In literature the practical application of proposed quantization methods is rarely discussed. 
To distinguish between the differences in applicability of quantization methods, we introduce four levels of quantization solutions, in decreasing order of practical applicability.
Our hope is that this will enable other authors to explore solutions for each level, and makes the comparison between methods more fair.
The axes for comparison are whether or not a method requires data, whether or not a method requires error backpropagation on the quantized model, and whether or not a method is generally applicable for any architecture or requires significant model re-working. 
We use the following definitions throughout the paper:

\begin{description}\setlength\itemsep{.05cm}
    \item[Level 1] 
    No data and no backpropagation required. Method works for any model. As simple as an API call that only looks at the model definition and weights.
    \item[Level 2] 
    Requires data but no backpropagation. Works for any model.
    The data is used e.g.\ to re-calibrate batch normalization statistics \cite{peters2018} or to compute layer-wise loss functions to improve quantization performance. However, no fine-tuning pipeline is required.
    \item[Level 3] 
    Requires data and backpropagation. Works for any model.
    Models can be quantized but need fine-tuning to reach acceptable performance. Often requires hyperparameter tuning for optimal performance.
    These methods require a full training pipeline (e.g.\ \cite{jacob2018cvpr, Zhou2017}).
    \item[Level 4] 
    Requires data and backpropagation. Only works for specific models.
    In this case, the network architecture needs non-trivial reworking, and/or the architecture needs to be trained from scratch with quantization in mind (e.g.\ \cite{pact2018, sheng2018, louizos2018relaxed}). Takes significant extra training-time and hyperparameter tuning to work. 
\end{description}




%% file: sections/related_work.tex
There are several works that describe quantization and improving networks for lower bit inference and deployment \cite{Gupta2015, Gysel2016, jacob2018cvpr, Zhou2017}. 
These methods all rely on fine-tuning, making them level 3 methods, whereas data-free quantization improves performance similarly without that requirement. 
Our method is complementary to these and can be applied as a pre-processing before quantization aware fine-tuning.

In a whitepaper, Krishnamoorthi \cite{krishnamoorthi}, introduces a level 1 `per-channel' quantization scheme, in which the weights of a convolutional weight tensor are quantized per output channel. A major drawback of this method is that it is not supported on all hardware, and that it creates unnecessary overhead in the computation due to the necessity of scale and offset values for each channel individually. 
We show that our method improves on per-channel quantization, while keeping a single set of scale and offset values for the whole weight tensor instead. 

Other methods to improve quantization need architecture changes or training with quantization in mind from the start \cite{achterhold2018variational, louizos2018relaxed, sheng2018,  ullrich2018, Zhou2016}. These methods are even more involved than doing quantization and fine-tuning. They also incur a relatively large overhead during training because of sampling and noisy optimization, and introduce extra hyperparameters to optimize. This makes them level 4 methods. 

Methods that binarize \cite{courbariaux2016,hubara2016, peters2018, rastegari2016} or ternarize \cite{Li2016} networks result in models with great inference efficiency as expensive multiplications and additions are replaced by bit-shift operations. 
However, quantizing models to binary often leads to strong performance degradation. Generally they need to be trained from scratch,  
making them level 4 methods.

Other approaches use low-bit floating point operations instead of integer operations, or other custom quantization implementations \cite{gudovskiy2017shiftcnn,  koster2017flex,  miyashita2016convlog, Zhou2017}. 
We do not consider such approaches as the hardware implementation is less efficient.

In concurrent work, Meller et al. \cite{samesame} also exploits the scale equivariance of the ReLU function to rescale weight channels and notice the biased error introduced by weight quantization \cite{biaswithbias}, leading to a method that resembles our data-free quantization approach. Stock et al. \cite{stock2018equinormalization} also use the scale equivariance property of the ReLU function, but use it for network optimization instead.


%% file: sections/motivation.tex
While many trained FP32 models can be quantized to INT8 without much loss in performance, some models exhibit a significant drop in performance after quantization (\cite{krishnamoorthi, sheng2018}). 
For example, when quantizing a trained MobileNetV2 \cite{mobilenetv2} model, Krishnamoorthi \cite{krishnamoorthi} reports a drop in top-1 accuracy from 70.9\% to 0.1\% on the ImageNet \cite{ILSVRC15} validation set. 
The author restores near original model performance by either applying per-channel quantization, fine-tuning or both. 

\subsection{Weight tensor channel ranges}\label{sec:weighttensorchannelranges}
The fact that per-channel quantization yields much better performance on MobileNetV2 than per-tensor quantization suggests that, in some layers, the weight distributions differ so strongly between output channels that the same set of quantization parameters cannot be used to quantize the full weight tensor effectively. 
For example, in the case where one channel has weights in the range $[-128, 128]$ and another channel has weights in the range $(-0.5, 0.5)$, the weights in the latter channel will all be quantized to $0$ when quantizing to 8-bits. 

\begin{figure}[t]
    \includegraphics[width=8cm, trim={.2cm .2cm .2cm .7cm}, clip]{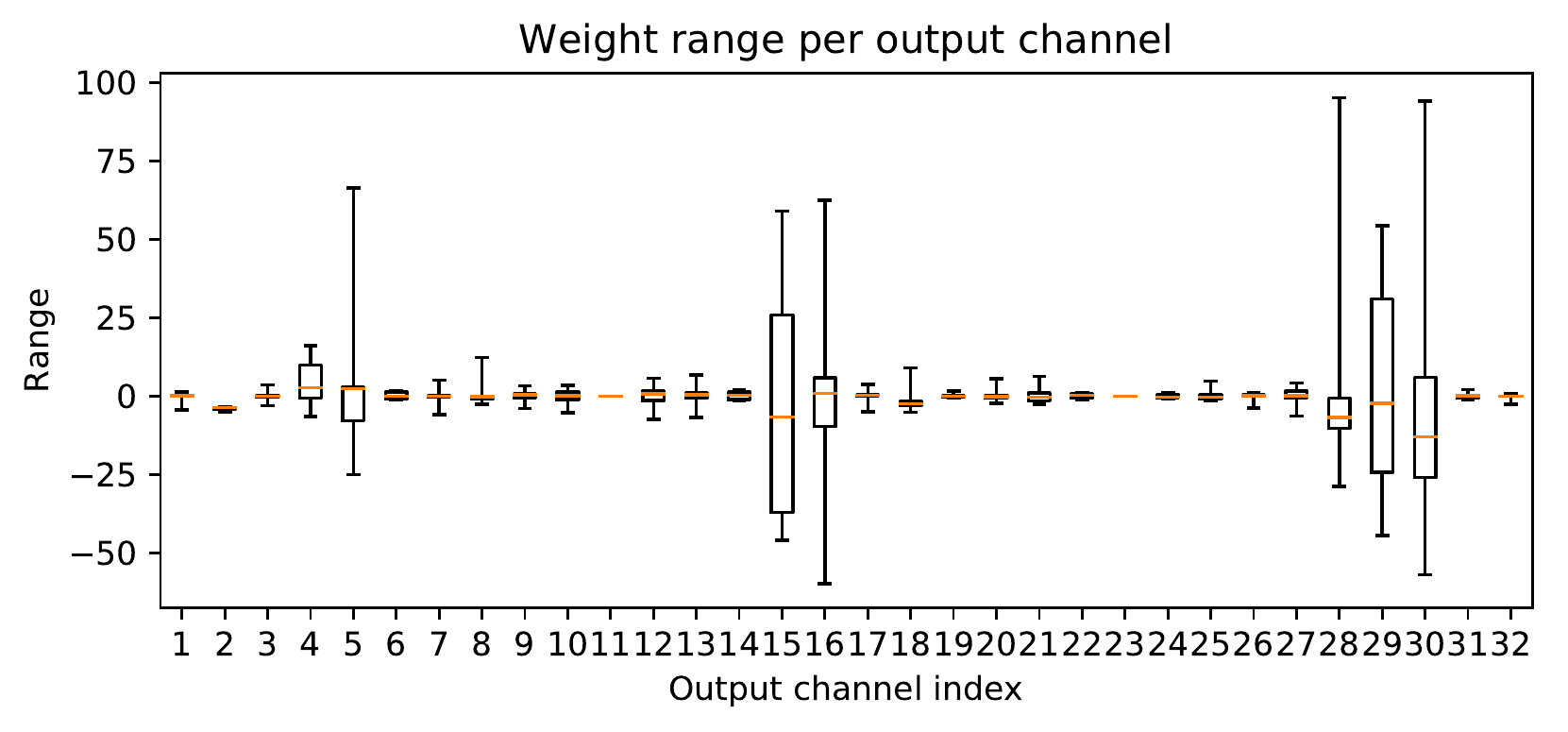}
    \centering
    \caption{Per (output) channel weight ranges of the first depthwise-separable layer in MobileNetV2. In the boxplot the min and max value, the 2nd and 3rd quartile and the median are plotted for each channel. This layer exhibits strong differences between channel weight ranges.}
\label{fig:mobilenet_channel_scales}
\end{figure}

Figure \ref{fig:mobilenet_channel_scales} shows that large differences in output channel weight ranges do indeed occur in a (trained) MobileNetV2 model. 
This figure shows the weight distribution of the output channel weights of the depthwise-separable layer in the model's first inverted residual block. 
Due to the strong differences between channel weight ranges that this layer exhibits, it cannot be quantized with reasonable accuracy for each channel. 
Several layers in the network suffer from this problem, making the overall model difficult to quantize.

We conjecture that performance of trained models after quantization can be improved by adjusting the weights for each output channel such that their ranges are more similar.
We provide a level 1 method to achieve this without changing the FP32 model output in section \ref{sec:equalization}.

\subsection{Biased quantization error}


A common assumption in literature (e.g.\ \cite{alvarez2016}) is that quantization error is unbiased and thus cancels out in a layer's output, ensuring that the mean of a layer's output does not change as a result of quantization.
However, as we will show in this section, the quantization error on the weights might introduce biased error on the corresponding outputs.
This shifts the input distribution of the next layer, which may cause unpredictable effects.


The biased error in a quantized layer's output unit $j$ can be computed empirically using $N$ input data points as:
\begin{align}
    \mathbb{E}[\yt_j - \y_j] &\approx
    \frac{1}{N}\sum_n (\Wt\xx_{n})_j - (\W\xx_{n})_j
    \label{eq:compute_output_bias}
\end{align}
where $\y_j$ and $\yt_j$ are the original outputs and the outputs generated using the quantized weight matrix, respectively.

\begin{figure}[t]
    \includegraphics[width=8cm, trim={.2cm .2cm .2cm .7cm}, clip]{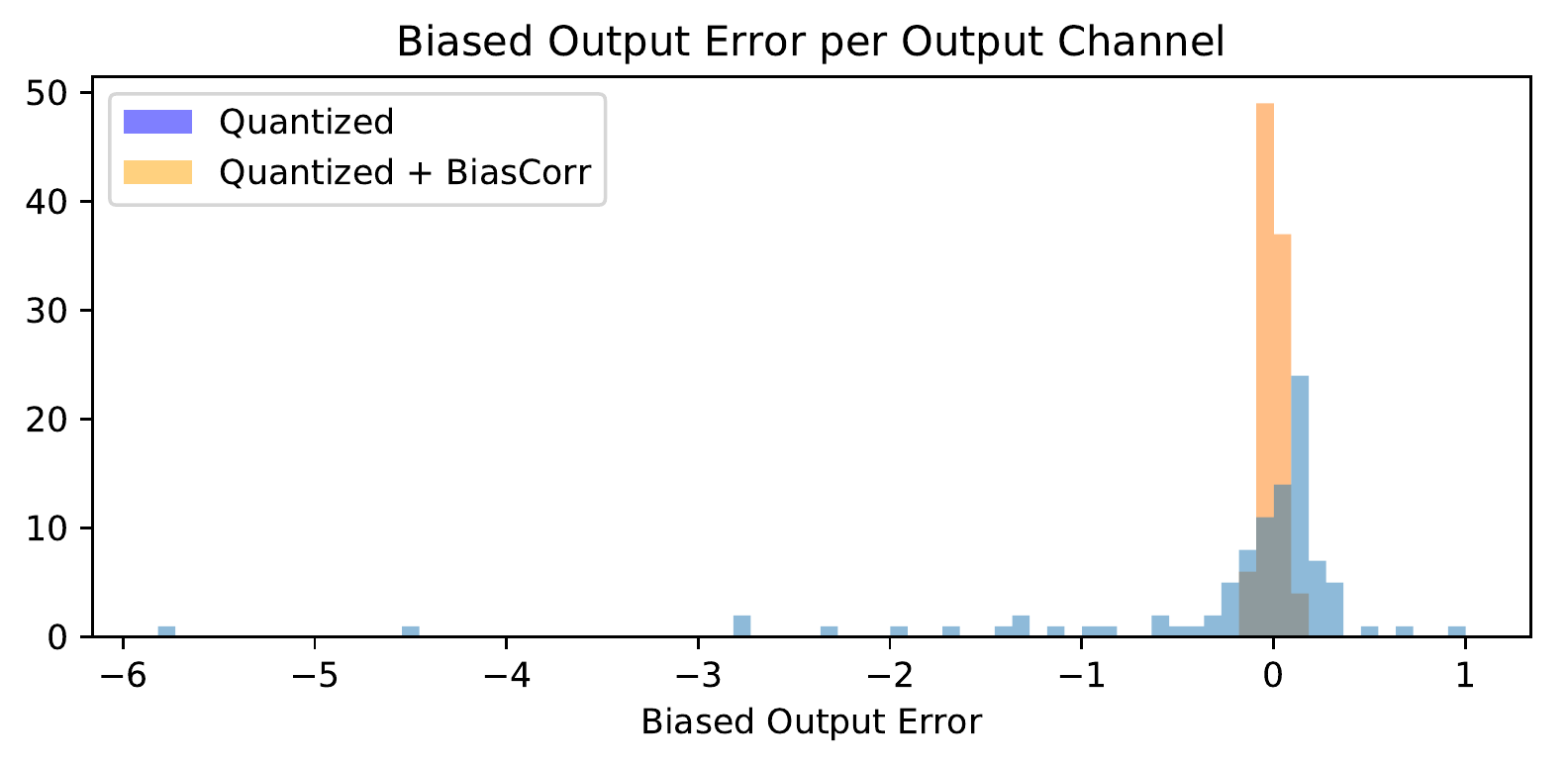}
    \centering
    \caption{Per-channel biased output error introduced by weight quantization of the second depthwise-separable layer in MobileNetV2, before (blue) and after (orange) bias correction.}
\label{fig:mobilenet_bias_fixed}
\end{figure}

Figure \ref{fig:mobilenet_bias_fixed} shows the biased error per channel of a depthwise-separable convolution layer in a trained MobileNetV2 model. 
From this plot it is clear that for many channels in the layer's output, the error introduced by weight quantization is biased, and influences the output statistics. 
Depthwise-separable layers are especially susceptible to this biased error effect as each output channel has only 9 corresponding weights.

Such a biased error on the outputs can be introduced in many settings, e.g.\ when weights or activations are clipped \cite{mishra2017}, or in non-quantization approaches, such as weight tensor factorization or channel pruning \cite{he2017, zhang2015}.

In section \ref{sec:bias_correction} we introduce a method to correct for this bias. Furthermore, we show that a model's batch normalization parameters can be used to compute the expected biased error on the output, yielding a level 1 method to fix the biased error introduced by quantization.



%% file: sections/method.tex
\begin{figure}[t]
\includegraphics[width=7cm]{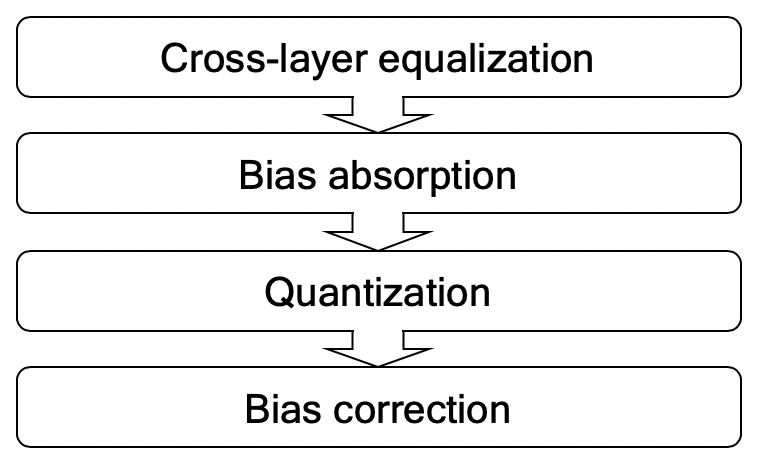}\vspace{-0.1cm}
\centering
\caption{Flow diagram of the proposed DFQ algorithm.}
\label{fig:flow_diagram}\vspace{-0.1cm}
\end{figure}

Our proposed data-free quantization method (DFQ) consists of three steps, on top of the normal quantization. The overall flow of the algorithm is shown in Figure \ref{fig:flow_diagram}.

\subsection{Cross-layer range equalization}\label{sec:equalization}

\paragraph{Positive scaling equivariance}\label{sec:positivescalingequivariance}
We observe that for a ReLU \cite{nair2010} activation function $f(\cdot)$ the following scaling equivariance property holds:
\begin{equation}
    f(sx) = sf(x)
\label{eq:equivariance}
\end{equation}
for any non-negative real number $s$. This follows from the definition of the ReLU: 
\begin{equation}
    \text{ReLU}(x)= 
    \begin{cases}
        x&\text{if } x > 0\\
        0&\text{if } x \leq 0.
    \end{cases}
\end{equation}

This equivariance also holds for the PreLU \cite{he2015} activation function.
More generally, the positive scaling equivariance can be relaxed to $f(sx) = s \hat{f}(x)$ for any piece-wise linear activation functions:
\begin{equation}
    f(x)=
    \begin{cases}
        a_1 x + b_1 &\text{if } x \leq c_1\\
        a_2 x + b_2 &\text{if } c_1 < x \leq c_2\\
        &\vdots \\
        a_n x + b_n &\text{if } c_{n-1} < x
    \end{cases}
\end{equation}
where $\hat{f}(\cdot)$ is parameterized as $\hat{a}_i = a_i$, $\hat{b}_i = b_i/s$ and $\hat{c}_i = c_i/s$. 
Note that contrary to equivariance defined in eq.\ \ref{eq:equivariance} we now also change the function $f(\cdot)$ into $\hat{f}(\cdot)$.

\subsubsection{Scaling equivariance in neural networks}
\begin{figure}[t]
\includegraphics[width=8cm]{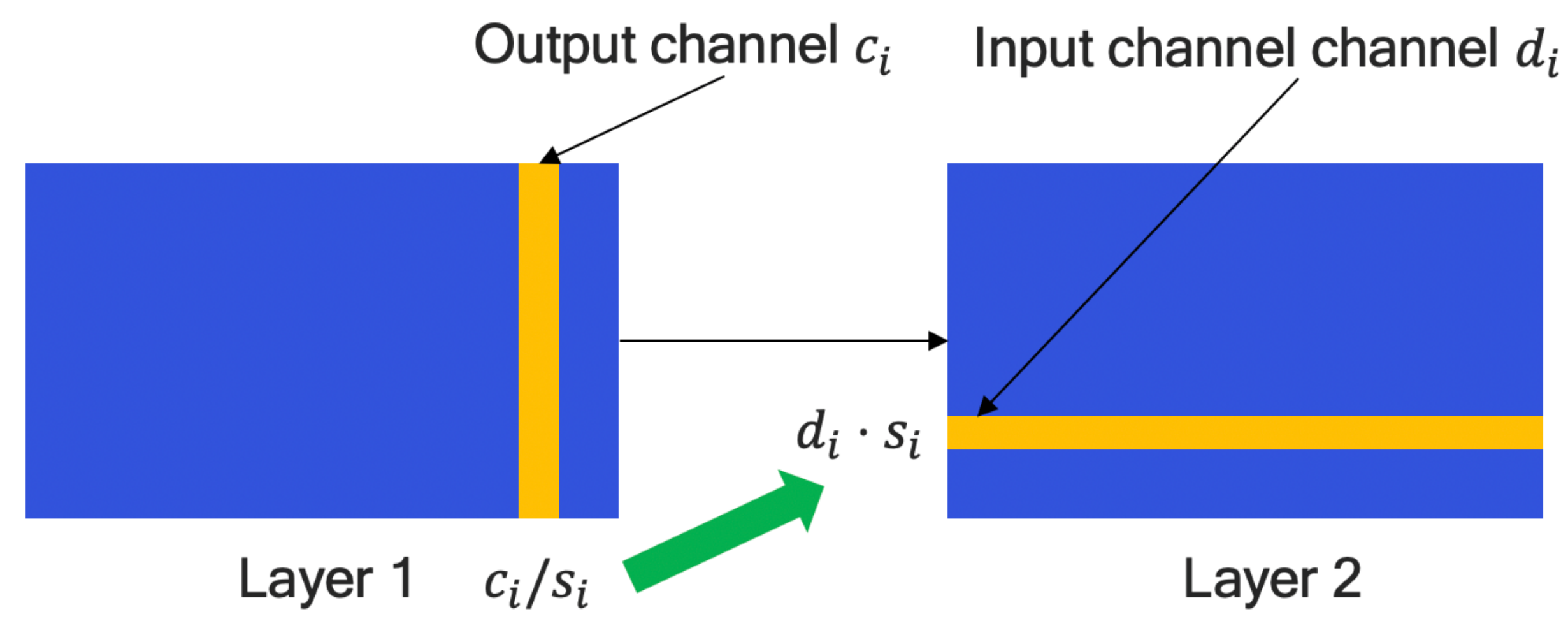}
\centering
\caption{Illustration of the rescaling for a single channel. If scaling factor $s_i$ scales $c_i$ in layer 1; we can instead factor it out and multiply $d_i$ in layer 2.}
\label{fig:layer_wise_rescaling}\vspace{-0.1cm}
\end{figure}

The positive scaling equivariance can be exploited in consecutive layers in neural networks. Given two layers, $\vec{h} = f(\mati{W}{1} \vec{x} + \veci{b}{1})$ and $\vec{y} = f(\mati{W}{2}\vec{h} + \veci{b}{2})$, through scaling equivariance we have that:
\begin{align}
    \vec{y} &= f(\mati{W}{2} f(\mati{W}{1}\vec{x}+\veci{b}{1}) + \veci{b}{2}) \\
        &= f(\mati{W}{2} \mat{S} \hat{f}(\mat{S^{-1}}\mati{W}{1} \vec{x} + \mat{S^{-1}} \veci{b}{1}) + \veci{b}{2}) \\
        &= f(\mati{\widehat{W}}{2} \hat{f}(\mati{\widehat{W}}{1} \vec{x} + \veci{\widehat{b}}{1}) + \veci{b}{2})
\end{align}
where $\mat{S}=diag(\vec{s})$ is a diagonal matrix with value $\mat{S}_{ii}$ denoting the scaling factor $\vec{s}_i$ for neuron $i$. 
This allows us to reparameterize our model with 
$\mati{\widehat{W}}{2} = \mati{W}{2} \mat{S}$, 
$\mati{\widehat{W}}{1} = \mat{S^{-1}} \mati{W}{1}$ and 
$\veci{\widehat{b}}{1} = \mat{S^{-1}} \veci{b}{1}$.
In case of CNNs the scaling will be per channel and broadcast accordingly over the spatial dimensions. The rescaling procedure is illustrated in Figure \ref{fig:layer_wise_rescaling}.

\subsubsection{Equalizing ranges over multiple layers}
We can exploit the rescaling and reparameterization of the model to make the model more robust to quantization. 
Ideally the ranges of each channel $i$ are equal to the total range of the weight tensor, meaning we use the best possible representative power per channel.
We define the precision of a channel as:
\begin{equation}
    \vec{\hat{p}}_i^{(1)} = \frac{\vec{\hat{r}}_i^{(1)}}{\hat{R}^{(1)}}
\end{equation}
where $\vec{\hat{r}}_i^{(1)}$ is the quantization range of channel $i$ in $\mati{\widehat{W}}{1}$ and $\hat{R}^{(1)}$ is the total range of $\mati{\widehat{W}}{1}$. 
We want to find $\mat{S}$ such that the total precision per channel is maximized:
\begin{equation}
    \max_{\mat{S}} \sum_i \vec{\hat{p}}_i^{(1)} \vec{\hat{p}}_i^{(2)}
    \label{eq:optimal_rescaling}
\end{equation}

In the case of symmetric quantization we have $\vec{\hat{r}}_i^{(1)} = 2\cdot \max_j |\mat{\widehat{W}_{ij}^{(1)}}|$ and $\hat{R}^{(1)}=2\cdot\max_{ij} |\mat{\widehat{W}_{ij}^{(1)}} |$. 
Solving eq.\ \ref{eq:optimal_rescaling} (see appendix \ref{app:equalization}) leads to the necessary condition: 
\begin{equation}
    \argmax_{j} \frac{1}{\vec{s}_j}\vec{r}_j^{(1)} = \argmax_{k} \vec{s}_k \vec{r}_k^{(2)}
\end{equation}
meaning the limiting channel defining the quantization range is given by $\argmax_{i} \vec{r}_i^{(1)} \vec{r}_i^{(2)}$. 
We can satisfy this condition by setting $\mat{S}$ such that:
\begin{equation}
    \vec{s}_i = \frac{1}{\vec{r}_i^{(2)}}\sqrt{\vec{r}_i^{(1)} \vec{r}_i^{(2)}}
\end{equation}
which results in $\forall i: \vec{r}_i^{(1)} = \vec{r}_i^{(2)}$. 
Thus the channel's ranges between both tensors are matched as closely as possible. 

When equalizing multiple layers at the same time, we iterate this process for pairs of layers that are connected to each other without input or output splits in between, until convergence.

\subsubsection{Absorbing high biases}
In case $\vec{s}_i < 1$ the equalization procedure increases bias $\vec{b}^{(1)}_i$. 
This could in turn increase the range of the activation quantization. 
In order to avoid big differences between per-channel ranges in the activations we introduce a procedure that absorbs high biases into the subsequent layer.

For a layer with ReLU function $r$, there is a non-negative vector $\vec{c}$ such that $r(\mat{W}\vec{x}+\vec{b}-\vec{c}) = r(\mat{W}\vec{x}+\vec{b}) - \vec{c}$.
The trivial solution $\vec{c}=\vec{0}$ holds for all $\vec{x}$.
However, depending on the distribution of $\vec{x}$ and the values of $\vec{W}$ and $\vec{b}$, there can be some values $\vec{c}_i>\vec{0}$ for which this equality holds for (almost) all $\vec{x}$.
Following the previous two layer example, these $\vec{c}_i$ can be absorbed from layer $1$ into layer $2$ as:
\begin{align}
    \vec{y} &= \mati{W}{2} \vec{h} + \veci{b}{2} \\
        &= \mati{W}{2} (r(\mati{W}{1} \vec{x} + \veci{b}{1}) + \vec{c}-\vec{c})  + \veci{b}{2} \\
        &= \mati{W}{2} (r(\mati{W}{1} \vec{x} + \veci{\hat{b}}{1}) + \vec{c})  + \veci{b}{2} \\
        &= \mati{W}{2} \vec{\hat{h}} + \veci{\hat{b}}{2}
\end{align}
where $\veci{\hat{b}}{2} = \mati{W}{2} \vec{c} + \veci{b}{2}$, $\vec{\hat{h}}=\vec{h} - \vec{c}$, and $\veci{\hat{b}}{1} = \veci{b}{1}-\vec{c}$. 

To find $\vec{c}$ without violating our data-free assumption we assume that the pre-bias activations are distributed normally with the batch normalization shift and scale parameters $\bbet$ and $\bgam$ as its mean and standard deviation.
We set $\vec{c}=\max(\vec{0},\bbet-3\bgam)$.
If $\vec{c}>0$, the equality introduced above will hold for the $99.865\%$ of values of $\vec{x}$ (those greater than $\vec{c}$) under the Gaussian assumption.
As we will show in section \ref{sec:exp_equalize}, this approximation does not harm the full precision performance significantly but helps for activation quantization.
Note that, in case data is available, the pre-bias distribution of $\vec{x}$ can be found empirically and used to set $\vec{c}$.


\subsection{Quantization bias correction}\label{sec:bias_correction}
As shown empirically in the motivation, quantization can introduce a biased error in the activations.
In this section we show how to correct for the bias in the error on the layer's output, and how we can use the network's batch normalization parameters to compute this bias without using data.

For a fully connected layer with weight tensor $\W$, quantized weights $\Wt$, and input activations $\xx$, we have $\yt = \Wt\xx$ and therefore $\yt = \y + \beps\xx$, where we define the quantization error $\beps=\Wt-\W$, $\y$ as the layer pre-activations of the FP32 model, and $\yt$ that layer with quantization error added.


If the expectation of the error for output $i$, $\mathbb{E}[\beps \textbf{x}]_i \neq 0$, then the mean of the output $i$ will change. 
This shift in distribution may lead to detrimental behavior in the following layers.
We can correct for this change by seeing that:
\begin{align}
    \mathbb{E}[\y]&=\mathbb{E}[\y]+\mathbb{E}[\beps\xx]-\mathbb{E}[\beps\xx]\\
    &=\mathbb{E}[\yt]-\mathbb{E}[\beps\xx].
\end{align}
Thus, subtracting the expected error on the output $\mathbb{E}\left[\beps\xx\right] = \beps \mathbb{E}\left[\xx\right]$ from the biased output $\yt$ ensures that the mean for each output unit is preserved.

For implementation, the expected error can be subtracted from the layer's bias parameter, since the expected error vector has the same shape as the layer's output.
This method easily extends to convolutional layers as described in Appendix \ref{app:biascorr}.

\subsubsection{Computing the expected input}
To compute the expected error of the output of a layer, the expected input to the layer $\mathbb{E}[\xx]$ is required. 
If a model does not use batch normalization, or there are no data-usage restrictions, $\mathbb{E}[\beps \xx]$ can be computed by comparing the activations before and after quantization. Appendix \ref{app:databias} explains this procedure in more detail.

\paragraph{Clipped normal distribution}
When the network includes batch normalization before a layer, we can use it to calculate $\mathbb{E}[\xx]$ for that layer without using data. We assume the pre-activation outputs of a layer are normally distributed, that batch normalization is applied before the activation function, and that the activation function is some form of the class of clipped linear activation functions (e.g.\ ReLU, ReLU6), which clips its input range to the range $[a, b]$ where $a < b$, and $b$ can be $\infty$.

Due to the centralization and normalization applied by batch normalization, the mean and standard deviation of the pre-activations are known: these are the batch normalization scale and shift parameters (henceforth referred to as $\bgam$ and $\bbet$ respectively).

To compute $\mathbb{E}[\xx]$ from the previous layer's batch normalization parameters, the mean and variance need to be adjusted to account for the activation function that follows the batch normalization layer. For this purpose we introduce the clipped normal distribution.
A clipped-normally distributed random variable $X$ is a normally distributed random variable with mean $\mu$ and variance $\sigma^2$, whose values are clipped to the range $[a, b]$
The mean and variance of the clipped normal distribution can be computed in closed form from $\mu$, $\sigma$, $a$ and $b$.
We present the mean of the clipped normal distribution for the ReLU activation function, i.e.\ $a=0$ and $b=\infty$ in this section, and refer the reader to Appendix \ref{app:clippednormal} for the closed form solution for the general clipped normal distribution.

The expected value for channel $c$ in $\xx$, $\mathbb{E}[\xx_c]$, which is the output of a layer with batch normalization parameters $\bbet_c$ and $\bgam_c$, followed by a ReLU activation function is:
\begin{align}
    \mathbb{E}[\xx_c] &= \mathbb{E}\left[\text{ReLU}\left( 
    \xx_c^{pre}\right)\right] \\
    &= \bgam_c\mathcal{N}\left( \frac{-\bbet_c}{\bgam_c} \right) + \bbet_c\left[1-\Phi\left( \frac{-\bbet_c}{\bgam_c} \right)\right]
\end{align}
where $\xx_c^{pre}$ is the pre-activation output for channel $c$, which is assumed to be normally distributed with mean $\bbet_c$ and variance $\bgam_c^2$, $\Phi(\cdot)$ is the normal CDF, and the notation $\mathcal{N}(x)$ is used to denote the normal $\mathcal{N}(x | 0,1)$ PDF.

%% file: sections/experiments.tex
In this section we present two sets of experiments to validate the performance of data-free quantization (DFQ).
We first show in section \ref{sec:ablation} the effect of the different aspects of DFQ and how they solve the problems observed earlier.
Then we show in section \ref{sec:other_models} how DFQ generalizes to other models and tasks, and sets a new state-of-the-art for level 1 quantization.

To allow comparison to previously published results, we use both weights and activations are quantized using 8-bit asymmetric, per-tensor quantization in all experiments.
Batch normalization is folded in the adjacent layer before quantization.
Weight quantization ranges are the min and max of the weight tensor.
Activation quantization ranges are set without data, by using the learned batch normalization shift and scale parameter vectors ${\bm\beta}$ and ${\bm\gamma}$ as follows:
We compute the activation range for channel $i$ as ${\bm\beta}_i\pm n\cdot{\bm\gamma}_i$ (with $n=6$), with the minimum clipped to 0 in case of ReLU activation.
We observed a wide range of $n$ can be used without significant performance difference. All experiments are done in Pytorch \cite{pytorch}. 
In appendix \ref{app:addres} we show additional experiments using short-term fine-tuning, symmetric quantization and per-channel quantization.

\subsection{Ablation study}\label{sec:ablation}
In this section we investigate the effect of our methods on a pre-trained MobileNetV2 \cite{mobilenetv2} model\footnote{We use the Pytorch implementation of MobileNetV2 provided by \url{https://github.com/tonylins/pytorch-mobilenet-v2}.}.
We validate the performance of the model on the ImageNet \cite{ILSVRC15} validation set.
We first investigate the effects of different parts of our approach through a set of ablation studies.

\subsubsection{Cross-layer equalization}\label{sec:exp_equalize}
In this section we investigate the effects of cross-layer equalization and high-bias folding. 
We compare these methods to two baselines: the original quantized model and the less hardware friendly per-channel quantization scheme. 

The models considered in this section employ residual connections \cite{heresidual}. 
For these networks we apply cross-layer equalization only to the layers within each residual block. 
MobileNetV2 uses ReLU6 activation functions, which clips activation ranges to $[0, 6]$.
To avoid ReLU6 requiring a different cut off per channel after applying the equalization procedure, we replace ReLU6 with regular ReLU.

The results of the equalization experiments are shown in Table \ref{tbl:exp1_scaling}. 
Similar to \cite{krishnamoorthi}, we observe that the model performance is close to random when quantizing the original model to INT8.
Further we note that replacing ReLU6 by ReLU does not significantly degrade the model performance.
Applying equalization brings us to within 2\% of FP32 performance, close to the performance of per-channel quantization.
We note that absorbing high biases results in a small drop in FP32 performance, but it boosts quantized performance by 1\% due to more precise activation quantization. 
Combining both methods improves performance over per-channel quantization, indicating the more efficient per-tensor quantization could be used instead.

\input{tables/exp1_scaling.tex}

\begin{figure}[t]
    \includegraphics[width=8cm, trim={.2cm .2cm .2cm .7cm}, clip]{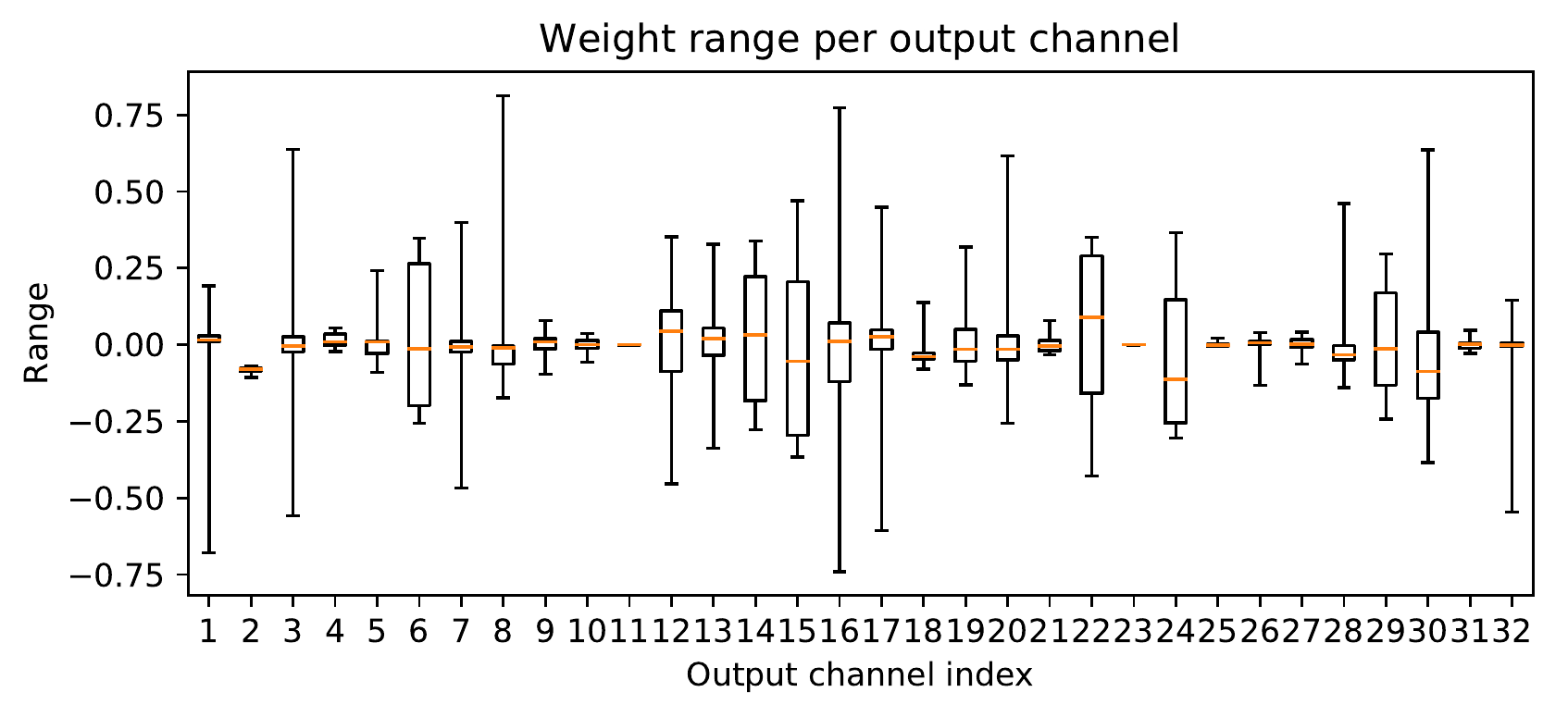}
    \centering
    \caption{Per (output) channel weight ranges of the first depthwise-separable layer in MobileNetV2 after equalization. In the boxplot the min and max value, the 2nd and 3rd quartile and the median are plotted for each channel. Most channels in this layer are now within similar ranges.}
\label{fig:mobilenet_channel_scales_fixed}
\end{figure}

To illustrate the effect of cross-layer equalization, we show the weight distributions per output channel of the depthwise-separable layer in the model’s first inverted residual block after applying the equalization in Figure \ref{fig:mobilenet_channel_scales_fixed}. 
We observe that most channels ranges are now similar and that the strong outliers from Figure \ref{fig:mobilenet_channel_scales} have been equalized. 
Note, there are still several channels which have all weight values close to zero. 
These channels convey little information and can be pruned from the network with hardly any loss in accuracy.

\subsubsection{Bias correction}
In this section we present results on bias correction for a quantized MobileNetV2 model.
We furthermore present results of bias correction in combination with a naive weight-clipping baseline, and combined with the cross-layer equalization approach.

The weight-clipping baseline serves two functions: 1) as a naive baseline to the cross-layer equalization approach, and 2) to show that bias correction can be employed in any setting where biased noise is introduced. 
Weight clipping solves the problem of large differences in ranges between channels by clipping large ranges to smaller ranges, but it introduces a strongly biased error. Weight clipping is applied by first folding the batch normalization parameters into a layer's weights, and then clipping all values to a certain range, in this case $[-15, 15]$. We tried multiple symmetric ranges, all provided similar results. For residual connections we calculate $\mathbb{E}[\xx]$ and $\text{Var}[\xx]$ based on the sum and variance of all input expectations, taking the input to be zero mean and unit variance.

To illustrate the effect of bias correction, Figure \ref{fig:mobilenet_bias_fixed} shows the per output channel biased error introduced by weight quantization.
The per-channel biases are obtained as described in eq.\ \ref{eq:compute_output_bias}.
This figure shows that applying bias correction reduces the bias in the error on the output of a layer to very close to 0 for most output channels.

\input{tables/exp1_bias.tex}

Results for the experiments described above for MobileNet V2 on the ImageNet validation set are shown in Table \ref{tbl:exp1_bias}.
Applying bias correction improves quantized model performance, indicating that a part of the problem of quantizing this model lies in the biased error that is introduced. However, bias correction on its own does not achieve near-floating point performance. The reason for this is most likely that the problem described in \ref{sec:motivation}.1 is more severe for this model. 
The experiments on weight-clipping show that bias correction can mitigate performance degradation due to biased error in non-quantized models as well as quantized models. Clipping without correction in the FP32 model introduces a 4.66\% loss in accuracy; bias correction reduces that loss to a mere 0.57\%. 
Furthermore, it shows that weight clipping combined with bias correction is a fairly strong baseline for quantizing MobileNet V2. 
Lastly, we show that bias correction improves results when combined with the cross-layer equalization and bias folding procedures. 
The combination of all methods is our data-free quantization (DFQ) method. The full DFQ approach achieves near-floating point performance with a reduction of 0.53\% top 1 accuracy relative to the FP32 baseline.






\subsection{Comparison to other methods and models}\label{sec:other_models}
In this section we show how DFQ generalizes to other popular computer vision tasks, namely semantic segmentation and object detection, and other model architectures such as MobileNetV1 \cite{mobilenetv1} and Resnet18 \cite{heresidual}. 
Afterwards we compare DFQ to methods in the literature, including more complex level 3 and 4 approaches. This set of models was chosen as they are efficient and likely to be used in mobile applications where 8-bit quantization is frequently used for power efficiency.

\subsubsection{Other tasks}

\paragraph{Semantic segmentation}
\input{tables/exp2_semseg.tex}
To demonstrate the generalization of our method to semantic segmentation we apply DFQ for DeeplabV3+ with a MobileNetV2 backend \cite{chen2018deeplab, mobilenetv2}, performance is evaluated on the Pascal VOC segmentation challenge \cite{everingham2015pascal}. For our experiments we use the publicly available Pytorch implementation\footnote{\url{https://github.com/jfzhang95/pytorch-deeplab-xception}}.

We show the results of this experiment in Table \ref{tbl:exp2_semseg}. As observed earlier for classification we notice a significant drop in performance when quantizing the original model which makes it almost unusable in practice. Applying DFQ recovers almost all performance degradation and achieves less than 1\% drop in mIOU compared to the full precision model. DFQ also outperforms the less hardware friendly per-channel quantization. 
To the best of our knowledge we are the first to publish quantization results on DeeplabV3+ as well as for semantic segmentation.

\paragraph{Object detection}
\input{tables/exp2_objdet.tex}
\input{tables/exp2_literature.tex}
To demonstrate the applicability of our method to object detection we apply DFQ for MobileNetV2 SSDLite \cite{mobilenetv2, liu2016ssd}, evaluated on the Pascal VOC object detection challenge \cite{everingham2015pascal}. In our experiments we use the publicly available Pytorch implementation of SSD\footnote{\url{https://github.com/qfgaohao/pytorch-ssd}}. 

The results are listed in Table \ref{tbl:exp2_objdet}. Similar to semantic segmentation we observe a significant drop in performance when quantizing the SSDLite model. Applying DFQ recovers almost all performance drop and achieves less than 1\% drop in mAP compared to the full precision model, again outperforming per-channel quantization.


\subsubsection{Comparison to other approaches}
In this section we compare DFQ to other approaches in literature. 
We compare our results to two other level 1 approaches, direct per-layer quantization as well as per-channel quantization \cite{krishnamoorthi}. 
In addition we also compare to multiple higher level approaches, namely quantization aware training \cite{jacob2018cvpr} as well as stochastic rounding and dynamic ranges \cite{Gupta2015, Gysel2016}, which are both level 3 approaches.
We also compare to two level 4 approaches based on relaxed quantization \cite{louizos2018relaxed}, which involve training a model from scratch and to quantization friendly separable convolutions \cite{sheng2018} that require a rework of the original MobileNet architecture.
The results are summarized in Table \ref{tbl:exp2_literature}.

For both MobileNetV1 and MobileNetV2 per-layer quantization results in an unusable model whereas DFQ stays close to full precision performance. DFQ also outperforms per-channel quantization as well as most level 3 and 4 approaches which require significant fine-tuning, training or even architecture changes.

On Resnet18 we maintain full precision performance for 8-bit fixed point quantization using DFQ. Some higher level approaches \cite{jacob2018cvpr, louizos2018relaxed} report slightly higher results than our baseline model, likely due to a better training procedure than used in the standard Pytorch Resnet18 model. 
Since 8-bit quantization is lossless we also compare 6-bit results. DFQ clearly outperforms traditional per-layer quantization but stays slightly below per-channel quantization and higher level approaches such as QT and RQ \cite{jacob2018cvpr, louizos2018relaxed}. 

Overall DFQ sets a new state-of-the-art for 8-bit fixed point quantization on several models and computer vision tasks. It is especially strong for mobile friendly architectures such as MobileNetV1 and MobileNetV2 which were previously hard to quantize.
Even though DFQ is an easy to use level 1 approach, we generally show competitive performance when comparing to more complex level 2-4 approaches.



%% file: tables/exp1_scaling.tex
\begin{table}[t]
    \centering
    \begin{tabular}{ l r r }
        \toprule
         Model          & FP32      & INT8 \\\midrule       
         Original model & 71.72\%   &  0.12\% \\            
         Replace ReLU6  & 71.70\%   &  0.11\% \\            
         + equalization    & 71.70\%   & 69.91\% \\            
         + absorbing bias & 71.57\%   & \textbf{70.92\%} \\   
         \midrule
         Per channel quantization & 71.72\% & 70.65\% \\
         \bottomrule \\ \vspace{-.6cm}
    \end{tabular}
    \caption{Top1 ImageNet validation results for MobileNetV2, evaluated at full precision and 8-bit integer quantized. Per-channel quantization is our own implementation of \cite{jacob2018cvpr} applied post-training.}
    \label{tbl:exp1_scaling}
\end{table}

%% file: tables/exp1_bias.tex

\begin{table}[t]
    \centering
    \begin{tabular}{l r r}
        \toprule
        Model & FP32 & INT8\\\midrule                                             
        Original Model          & 71.72\%           & 0.12\%\\                    
        Bias Corr               & 71.72\%           & \textbf{52.02}\%\\\midrule  
        Clip @ 15               & 67.06\%           & 2.55\%\\                   
        + Bias Corr   & \textbf{71.15}\%  & \textbf{70.43}\%\\\midrule            
        Rescaling + Bias Absorption& 71.57\%           & 70.92\%\\                   
        + Bias Corr   & 71.57\%           & \textbf{71.19}\%\\                    
        \bottomrule\\ \vspace{-.6cm}
    \end{tabular}
    \caption{Top1 ImageNet validation results for MobileNetV2, evaluated at full precision and 8-bit integer quantized. Bold results show the best result for each column in each cell.}
    \label{tbl:exp1_bias}
\end{table}

%% file: tables/exp2_semseg.tex
\begin{table}[t]
    \centering
    \begin{tabular}{ l r r }
        \toprule
         Model & FP32 & INT8 \\ \midrule
         Original model & 72.94 & 41.40 \\  
         DFQ (ours) & 72.45 & \textbf{72.33} \\
         \midrule
         Per-channel quantization & 72.94 & 71.44 \\
         \bottomrule \\ \vspace{-.6cm}
    \end{tabular}
    \caption{DeeplabV3+ (MobileNetV2 backend) on Pascal VOC segmentation challenge. Mean intersection over union (mIOU) evaluated at full precision and 8-bit integer quantized. Per-channel quantization is our own implementation of \cite{jacob2018cvpr} applied post-training.}
    \label{tbl:exp2_semseg}
\end{table}

%% file: tables/exp2_objdet.tex
\begin{table}[t]
    \centering
    \begin{tabular}{ l r r }
        \toprule
         Model & FP32 & INT8 \\ \midrule  
         Original model & 68.47 & 10.63 \\
         DFQ (ours) & 68.56 & \textbf{67.91} \\ 
         \midrule
         Per-channel quantization & 68.47 & 67.52 \\ 
         \bottomrule \\ \vspace{-.6cm}
    \end{tabular}
    \caption{MobileNetV2 SSD-lite on Pascal VOC object detection challange. Mean average precision (mAP) evaluated at full precision and 8-bit integer quantized. Per-channel quantization is our own implementation of \cite{jacob2018cvpr} applied post-training.}
    \label{tbl:exp2_objdet} 
\end{table}

%% file: tables/exp2_literature.tex
\begin{table*}[t]
    \centering
    \begin{tabular}{ l c c c r r r r r r r}
        \toprule 
        & $\sim$D & $\sim$BP & $\sim$AC & \multicolumn{2}{c}{MobileNetV2} & \multicolumn{2}{c}{MobileNetV1} & \multicolumn{3}{c}{ResNet18}\\
        & & & & FP32 & INT8 & FP32 & INT8 & FP32 & INT8 & INT6 \\ \midrule
        DFQ (ours) & \good & \good & \good & 71.7\% & \textbf{71.2\%} & 70.8\% & \textbf{70.5\%} & 69.7\% & \textbf{69.7\%} & {66.3\%}\\  
        Per-layer \cite{krishnamoorthi} & \good & \good & \good & 71.9\% & 0.1\% & 70.9\% & 0.1\% & 69.7\% & 69.2\%$^*$ & 63.8\%$^*$ \\ 
        Per-channel \cite{krishnamoorthi} & \good & \good & \good & 71.9\% & 69.7\% & 70.9\% & 70.3\% & 69.7\% & 69.6\%$^*$ & \textbf{67.5\%}$^*$ \\ 
        \midrule
        QT \cite{jacob2018cvpr} \^\  & \bad & \bad & \good & 71.9\% & 70.9\% & 70.9\% & 70.0\% & - & \textbf{70.3\%}$^\dagger$ & 67.3\%$^\dagger$ \\  
        SR+DR$^\dagger$  & \bad & \bad & \good & - & - & - & \textbf{71.3}\% & - & 68.2\% & 59.3\% \\
        QMN \cite{sheng2018}  & \bad & \bad & \bad & - & - & 70.8\% & 68.0\% & - & - & - \\  
        RQ \cite{louizos2018relaxed}  & \bad & \bad & \bad & - & - & - & 70.4\% & - & 69.9\% & \textbf{68.6\%} \\
        \bottomrule \\ \vspace{-.6cm}
    \end{tabular}
    \caption{Top1 ImageNet validation results for different models and quantization approaches. 
    The top half compares level 1 approaches ($\sim$D: data free, $\sim$BP: backpropagation-free, $\sim$AC: Architecture change free) whereas in the second half we also compare to higher level approaches in literature. 
    Results with $^*$ indicates our own implementation since results are not provided,  \^\ results provided by \cite{krishnamoorthi} and $^\dagger$ results from table 2 in \cite{louizos2018relaxed}. 
    }
    \label{tbl:exp2_literature} \vspace{-.2cm}
\end{table*}

%% file: sections/conclusion.tex
In this work, we introduced DFQ, a data-free quantization method that significantly helps quantized model performance without the need for data, fine-tuning or hyper-parameter optimization. 
The method can be applied to many common computer vision architectures with a straight-forward API call. 
This is crucial for many practical applications where engineers want to deploy deep learning models trained in FP32 to INT8 hardware without much effort. 
Results are presented for common computer vision tasks like image classification, semantic segmentation and object detection. 
We show that our method compares favorably to per-channel quantization \cite{krishnamoorthi}, meaning that instead the more efficient per-tensor quantization can be employed in practice. 
DFQ achieves near original model accuracy for almost every model we tested, and even competes with more complicated training based methods.

Further we introduced a set of quantization levels to facilitate the discussion on the applicability of quantization methods. 
There is a difference in how easy a method is to use for generating a quantized model, which is a significant part of the impact potential of a quantization method in real world applications. 
We hope that the quantization levels and methods introduced in this paper will contribute to both future research and practical deployment of quantized deep learning models. 



%% file: sections/appendix-cle.tex
Consider two fully-connected layers with weight matrices $\mat{W}^{(1)}$ and $\mat{W}^{(2)}$, that we scale as in \ref{sec:positivescalingequivariance}. We investigate the problem of optimizing the quantization ranges by rescaling the weight matrices by $\mat{S} = diag(\vec{s})$, where $\vec{s} > 0$, such that $\mat{\widehat{W}}^{(1)} = \mat{S}^{-1} \mat{W}^{(1)} $ and $\mat{\widehat{W}}^{(2)} = \mat{W}^{(2)} \mat{S}$ the weight matrices after rescaling. We investigate the case of symmetric quantization, which also gives good results in practice for asymmetric quantization. We denote
\begin{align}
\vec{\hat{r}}^{(1)} &= 2 \cdot\max_j |\mat{S^{-1}}\mat{W}^{(1)}_{ij}| = \mat{S^{-1}}\vec{r}^{(1)}\\
\vec{\hat{r}}^{(2)} &= 2 \cdot \max_i |\mat{W}^{(2)}_{ij}\mat{S}| = \vec{r}^{(2)}\mat{S} \\
\hat{R}^{(k)} &= max_i (\vec{r}_i^{(k)})
\end{align}
where $\vec{\hat{r}}^{(k)}$ are the per-channel weight ranges that are scaled by $\mat{S}$, $\hat{R}^{(k)}$ the range for the scaled weight matrix $\mat{\widehat{W}}^{(k)}$ and ${\vec{r}}^{(k)}$ are the original unscaled ranges.

Using this in our optimization goal of eq.\ \ref{eq:optimal_rescaling} leads to
\begin{align}
    &\max_{\mat{S}} \sum_i \vec{\hat{p}}_i^{(1)} \vec{\hat{p}}_i^{(2)} = \max_{\mat{S}} \sum_i \frac{\vec{\hat{r}}_i^{(1)} \vec{\hat{r}}_i^{(2)}}{\hat{R}^{(1)}\hat{R}^{(2)}} \\
    &= \max_{\mat{S}} \sum_i \frac{\frac{1}{\vec{s}_i}\vec{r}_i^{(1)} \cdot \vec{s}_i \vec{r}_i^{(2)} }
        {\max_j (\frac{1}{\vec{s}_j} \vec{r}_j^{(1)}) \cdot \max_k (\vec{s}_k \vec{r}_k^{(2)})} \\
    &= \sum_i \vec{r}_i^{(1)} \vec{r}_i^{(2)} \max_{\mat{S}} \frac{1}{\max_j (\frac{1}{\vec{s}_j} \vec{r}_j^{(1)}) \cdot \max_k (\vec{s}_k \vec{r}_k^{(2)})}.
\end{align}
We observe that the specific scaling $\vec{s}_i$ of each channel cancels out as long as they do not increase $R$, the range of the full weight matrix. 
We can reformulate the above to
\begin{equation}
    \min_{\mat{S}} \left( \max_j \left(\frac{1}{\vec{s}_j} \vec{r}_j^{(1)}\right) \cdot \max_k (\vec{s}_k \vec{r}_k^{(2)}) \right)
    \label{eq:opt_app}
\end{equation}
which is minimal iff
\begin{equation}
    \argmax_{j} \frac{1}{\vec{s}_j} \vec{r}_j^{(1)} = \argmax_{k} \vec{s}_k \vec{r}_k^{(2)}
    \label{eq:opt_channel}
\end{equation}
By contradiction, if $j \neq k$ there is a small positive $\epsilon$ such that $\vec{s}'_k = \vec{s}_k - \epsilon$
which will decrease $\max_k \vec{s}_k\vec{r}_k^{(2)}$ by $\epsilon \vec{r}_k^{(2)}$ without affecting $\max_j \frac{1}{\vec{s}_j} \vec{r}_j^{(1)}$. Therefore such a solution would not be optimal for eq. \ref{eq:opt_app}.

The condition from eq.\ \ref{eq:opt_channel} implies there is a limiting channel $i = \argmax_{i} \vec{r}_i^{(1)} \vec{r}_i^{(2)}$ which defines the quantization range of both weight matrices $\mat{\widehat{W}}^{(1)}$ and $\mat{\widehat{W}}^{(2)}$. 
However, our optimization goal is not affected by the choice of the other $\vec{s}_j$ given the resulting $\vec{\hat{r}}_j^{(1)}$ and $\vec{\hat{r}}_j^{(2)}$ are smaller than or equal to $\vec{\hat{r}}_i^{(1)}$ and $\vec{\hat{r}}_i^{(2)}$, respectively.
To break the ties of solutions we decide to set $\forall i: \vec{r}_i^{(1)} = \vec{r}_i^{(2)}$. 
Thus the channel's ranges between both tensors are matched as closely as possible and the introduced quantization error is spread equally among both weight tensors.
This results in our final rescaling factor
\begin{equation}
    \vec{s}_i = \frac{1}{\vec{r}_i^{(2)}}\sqrt{\vec{r}_i^{(1)} \vec{r}_i^{(2)}}
\end{equation}
which satisfies our necessary condition from eq.\ \ref{eq:opt_channel} and ensures that $\forall i: \vec{r}_i^{(1)} = \vec{r}_i^{(2)}$.

%% file: sections/appendix-biascorr.tex
Similarly to fully connected layers we can compute $\beps$ from $\W$ and $\Wt$, it becomes a constant and we have that $\mathbb{E}\left[\beps*\xx\right]=\beps*\mathbb{E}[\xx]$.
Expanding this yields:
\begin{align}
    \left[\beps * \mathbb{E}[\xx]\right]_{c_oij} &= \sum_{c_imn}\mathbb{E}[\xx_{c_i,i-m,j-n}]\beps_{c_oc_imn} \\
    &= \sum_{c_i}\left[\mathbb{E}[\xx_{c_i}]\sum_{mn}\beps_{c_oc_imn}\right]
\end{align}
where we assume that the expected value of each input channel is the same for all spatial dimensions in the input channel. Since the value of $\left[\beps * \mathbb{E}[\xx]\right]_{c_oij}$ does not depend on the spatial dimensions $i$ and $j$, the expected error is the same for the full output channel and can be folded into the layer's bias parameter.

%% file: sections/appendix-clipnorm.tex
Given a normally distributed random variable $X$ with mean $\mu$ and variance $\sigma^2$, and a clipped-linear function $f(\cdot)$ that clips its argument to the range $[a, b]$, s.t. $a < b$, the mean and variance of $f(X)$ can be determined using the standard rules of computing the mean and variance of a function:
\begin{align}
    \muclip &= \int_{-\infty}^{\infty} f(x)p(x)dx\\
    \sigclip^2 &= \int_{-\infty}^{\infty}(f(x)-\muclip)^2p(x)dx
\end{align}
where we define $p(x)=\mathcal{N}(x\mid\mu, \sigma)$, $\muclip=\mathbb{E}[f(X)]$ and $\sigclip^2 = Var[f(X)]$.

\subsection{Mean of Clipped Normal Distribution}
Using the fact that $f(x)$ is constant if $x\not\in[a,b]$ we have that:
\begin{align}
    \muclip &= \int_{-\infty}^{\infty} f(x)p(x)dx\\
    &= a\int_{-\infty}^a p(x)dx + \int_a^bxp(x)dx + b\int_b^{\infty}p(x)dx
\end{align}
The first and last term can be computed as $a\Phi(\alpha)$ and $b(1-\Phi(\beta))$ respectively, where we define $\alpha=\frac{a-\mu}{\sigma}$,  $\beta=\frac{b-\mu}{\sigma}$, and $\Phi(x)=CDF(x \mid 0, 1)$, the normal CDF with zero mean and unit variance.

The integral over the linear part of $f(\cdot)$ can be computed as:
\begin{align}
    \int_a^bxp(x)dx &= C\int_a^bxe^{-\frac{1}{2\sigma^2}(x-\mu)^2}dx \\
    &= \left.-C\sigma^2e^{-\frac{1}{2\sigma^2}(x-\mu)^2}\right\rvert_a^b + \mu(\Phi(\beta) - \Phi(\alpha))\\
    &= \sigma\left(\phi(\alpha)-\phi(\beta)\right) + \mu(\Phi(\beta) - \Phi(\alpha))
\end{align}
where we define $\phi(\cdot)=\mathcal{N}(\cdot\mid0, 1)$, i.e. the standard normal pdf and $C=\frac{1}{\sigma\sqrt{2\pi}}$ is the normalization constant for a normal distribution with variance $\sigma^2$, thus
\begin{equation}
    \begin{aligned}
    \muclip =& \sigma\left(\phi(\alpha)-\phi(\beta)\right) + \mu(\Phi(\beta) - \Phi(\alpha)) 
    \\&+ a\Phi(\alpha) + b(1-\Phi(\beta)).
    \end{aligned}
\end{equation}

\subsection{Variance of Clipped Normal Distribution}
We again exploit the fact that $f(x)$ is constant if $x\not\in[a,b]$:
\begin{align}
    \sigclip^2 &=\int_{-\infty}^{\infty} (f(x) -\muclip)^2p(x)dx \\
    &\begin{aligned}
        &=\int_{-\infty}^a(a-\muclip)^2p(x)dx + \\
        &+ \int_a^b(x-\muclip)^2p(x)dx + \\
        &+\int_b^\infty (b-\muclip)^2p(x)dx
    \end{aligned}
\end{align}
The first and last term can be solved as $(a-\muclip)^2\Phi(\alpha)$ and $(b-\muclip)^2(1-\Phi(\beta))$ respectively.

The second term can be decomposed as follows:
\begin{align}
    \int_a^b(x-\muclip)^2p(x)dx &=
    \int_a^b(x^2-2x\muclip+\muclip^2)p(x)dx \\
    &\begin{aligned}
    &= \int_a^b x^2p(x)dx \\
    &+ Z(\muclip^2 - 2\muclip\mutrunc)
    \end{aligned}
\end{align}
where we use the result from the previous subsection and define $Z=\Phi(\beta)-\Phi(\alpha)$, and where $\mutrunc=\frac{1}{Z}\int_a^b x\mathcal{N}(x\mid \mu, \sigma^2)=\mu+\sigma (\phi(\alpha) - \phi(\beta))/Z$ is the mean of the truncated normal distribution.

Evaluating the first term yields:
\begin{equation}
    \begin{aligned}
        \int_a^bx^2p(x)dx &= Z(\mu^2+\sigma^2) \\
        &+ \sigma(a\phi(\alpha) - b\phi(\beta)) \\
        &+ \sigma\mu(\phi(\alpha) - \phi(\beta))
    \end{aligned}
\end{equation}

This results in:
\begin{equation}
    \begin{aligned}
        Var[f(X)] &= Z(\mu^2+\sigma^2+ \muclip^2 - 2\muclip\mu) \\
        &+\sigma(a\phi(\alpha)-b\phi(\beta)) \\
        &+ \sigma(\mu-2\muclip)(\phi(\alpha)-\phi(\beta)) \\
        &+ (a-\muclip)^2\Phi(\alpha) \\
        &+ (b-\muclip)^2(1-\Phi(\beta))
    \end{aligned}
\end{equation}

%% file: sections/appendix-empirical-quant-bias-correction.tex

\begin{table}[t]
    \centering
    \begin{tabular}{ l r r }
        \toprule
         Model               & CLE+BA   & Clip@15 \\ \midrule
         No BiasCorr         & 70.92\%  & 2.55\%  \\
         Analytic BiasCorr   & 71.19\%  & 70.43\% \\
         Empirical BiasCorr & 71.15\%  & 69.85\% \\
         \bottomrule \\ \vspace{-.7cm}
    \end{tabular}
    \caption{Top1 ImageNet validation results for MobileNetV2 with weights and activations quantized to INT8. Comparing analytic and empirical bias correction combined with cross-layer equalization (CLE), bias absorption (BA) and clipping.}
    \label{tbl:exp1_databias}
\end{table}

If a network does not use batch normalization, or does not use batch normalization in all layers, a representative dataset can be used to compute the difference between pre-activation means before and after quantization. 
We then subtract this difference from the quantized model's pre-activations.
This procedure can be run with unlabeled data. 
The procedure should be run after BatchNorm folding and cross-layer range equalization. 
Clipping should be applied in the quantized network, but not in the floating point network. 
Since the activation function and the quantization operation are fused, this procedure is run on a network with quantized weights only. 
However, after this procedure is applied activations can be quantized as well.
We bias correct a layer only after all the layers feeding into it have been bias-corrected.
The procedure is as follows:

\begin{enumerate}
    \item Run $N$ examples through the FP32 model and collect for each layer the per-channel pre-activation means $\mathbb{E}[\y]$.
    \item For each layer $L$ in the quantized model:
    \begin{itemize}
        \item Collect the per-channel pre-activation means $\mathbb{E}[\yt]$ of layer $L$ for the same $N$ examples as in step 1.
        \item Compute the per-channel biased quantization error $\mathbb{E}[\beps]=\mathbb{E}[\yt]-\mathbb{E}[\y]$.
        \item Subtract $\mathbb{E}[\beps]$ from the layer's bias parameter.
    \end{itemize}
\end{enumerate}

In Table \ref{tbl:exp1_databias} we compare this empirical bias correction procedure with the analytic bias correction introduced in section \ref{sec:bias_correction}. We observe that both approaches leads to similar results.


%% file: sections/appendix-additional-results.tex
\paragraph{Combination with fine-tuning}
The focus of our method is data-free quantization (level 1). However, our method can also be used as a pre-processing before quantization aware fine-tuning.
To demonstrate this we used DFQ together with short-term quantization aware fine-tuning \cite{krishnamoorthi}. 
After just 1 epoch of quantization aware fine-tuning MobileNet V2, accuracy increases from 71.19\% to 71.42\%, almost recovering the FP32 performance (71.72\%).

\paragraph{Symmetric vs asymmetric quantization}
In our experimental section all our experiments were performed with asymmetric quantization, since this is commonly used in literature. Here we also compare to symmetric quantization. Symmetric quantization does not use an offset, which eliminates several cross terms in the calculations done on hardware compared to asymmetric quantization. This makes symmetric quantization more efficient on some hardware at the expense of losing some expressive power.

In Table \ref{tab:symmetric} we compare symmetric and asymmetric quantization in combination with DFQ. For all three models the advantage of asymmetric quantization is almost negligible. 
We noticed that cross-layer equalization is effective at removing outliers, resulting in weight distributions are often close to symmetric.

\begin{table}[t]
\centering
    \begin{tabular}{ l r r }
        \toprule
         Model                  & Symmetric   & Asymmetric\\ \midrule 
         MobileNet V1           & 70.32\%   & 70.51\% \\
         MobileNet V2           & 71.15\%   & 71.19\% \\
         Resnet18               & 69.50\%   & 69.62\% \\
         \bottomrule  \\ \vspace{-.7cm}
    \end{tabular} 
    \caption{Top1 ImageNet validation results for MobileNetV2 after applying DFQ. Weights and activations quantized using symmetric and asymmetric 8-bit integer quantization.}
    \label{tab:symmetric}
\end{table}

\paragraph{DFQ combined with per-channel quantization}
In our experiments we focused on per-tensor quantization since the more recent per-channel quantization is not efficiently supported on all hardware. For hardware that does support it, we analyze the effect of DFQ in combination with per-channel quantization.

In Table \ref{tab:per_channel} we show the results of the different components of DFQ in combination with per-channel quantization. 
We notice that each individual component, cross-layer equalization, bias absorption and bias correction, incremental improve over per-channel quantization and reduce the total quantization error from 1.07\% to only 0.39\%.

\begin{table}[t]
\centering
    \begin{tabular}{ l r r }
        \toprule
         Model                  & No BiasCorr   & BiasCorr\\ \midrule 
         Original model         & 70.65\%   & 70.80\% \\
         CLE                    & 70.93\%   & 71.30\% \\
         CLE+BA  & 71.03\%   & \textbf{71.33\%} \\
         \bottomrule \\ \vspace{-.7cm}
    \end{tabular}
        \caption{Top1 ImageNet validation results for MobileNetV2 using 8-bit per-channel quantization. Activations are quantized per tensor. Showing the effect of cross-layer equalization (CLE) and bias absorption (BA) in combination with bias correction.}
    \label{tab:per_channel}
\end{table}

%% file: main.bbl
\begin{thebibliography}{10}\itemsep=-1pt

\bibitem{achterhold2018variational}
Jan Achterhold, Jan~Mathias Koehler, Anke Schmeink, and Tim Genewein.
\newblock Variational network quantization.
\newblock In {\em International Conference on Learning Representations (ICLR)},
  2018.

\bibitem{alvarez2016}
Raziel Alvarez, Rohit Prabhavalkar, and Anton Bakhtin.
\newblock On the efficient representation and execution of deep acoustic
  models.
\newblock In {\em The Annual Conference of the International Speech
  Communication Association (Interspeech)}, 2016.

\bibitem{chen2018deeplab}
Liang-Chieh Chen, Yukun Zhu, George Papandreou, Florian Schroff, and Hartwig
  Adam.
\newblock Encoder-decoder with atrous separable convolution for semantic image
  segmentation.
\newblock In {\em The European Conference on Computer Vision (ECCV)}, September
  2018.

\bibitem{pact2018}
Jungwook Choi, Zhuo Wang, Swagath Venkataramani, Pierce~I{-}Jen Chuang,
  Vijayalakshmi Srinivasan, and Kailash Gopalakrishnan.
\newblock {PACT:} parameterized clipping activation for quantized neural
  networks.
\newblock {\em arXiv preprint arxiv:805.06085}, 2018.

\bibitem{courbariaux2016}
Matthieu Courbariaux, Yoshua Bengio, and Jean-Pierre David.
\newblock Binaryconnect: Training deep neural networks with binary weights
  during propagations.
\newblock In {\em Proceedings of the 28th International Conference on Neural
  Information Processing Systems - Volume 2}, NIPS'15, pages 3123--3131,
  Cambridge, MA, USA, 2015. MIT Press.

\bibitem{everingham2015pascal}
Mark Everingham, {S. M. Ali} Eslami, Luc {Van Gool}, {Christopher K. I.}
  Williams, John Winn, and Andrew Zisserman.
\newblock The pascal visual object classes challenge: A retrospective.
\newblock {\em International Journal of Computer Vision}, 111(1):98--136, 1
  2015.

\bibitem{biaswithbias}
Alexander Finkelstein, Uri Almog, and Mark Grobman.
\newblock Fighting quantization bias with bias.
\newblock {\em arXiv preprint arxiv:1906.03193}, 2019.

\bibitem{gudovskiy2017shiftcnn}
Denis~A. Gudovskiy and Luca Rigazio.
\newblock Shiftcnn: Generalized low-precision architecture for inference of
  convolutional neural networks.
\newblock {\em arXiv preprint arxiv:1706.02393}, 2017.

\bibitem{Gupta2015}
Suyog Gupta, Ankur Agrawal, Kailash Gopalakrishnan, and Pritish Narayanan.
\newblock Deep learning with limited numerical precision.
\newblock In {\em Proceedings of the 32nd International Conference on Machine
  Learning, {ICML} 2015, Lille, France, 6-11 July 2015}, pages 1737--1746,
  2015.

\bibitem{Gysel2016}
Philipp Gysel, Jon~J. Pimentel, Mohammad Motamedi, and Soheil Ghiasi.
\newblock Ristretto: {A} framework for empirical study of resource-efficient
  inference in convolutional neural networks.
\newblock {\em {IEEE} Trans. Neural Netw. Learning Syst.}, 29(11):5784--5789,
  2018.

\bibitem{he2015}
Kaiming He, Xiangyu Zhang, Shaoqing Ren, and Jian Sun.
\newblock Delving deep into rectifiers: Surpassing human-level performance on
  imagenet classification.
\newblock In {\em 2015 {IEEE} International Conference on Computer Vision,
  {ICCV} 2015, Santiago, Chile, December 7-13, 2015}, pages 1026--1034, 2015.

\bibitem{heresidual}
Kaiming He, Xiangyu Zhang, Shaoqing Ren, and Jian Sun.
\newblock Deep residual learning for image recognition.
\newblock In {\em 2016 {IEEE} Conference on Computer Vision and Pattern
  Recognition, {CVPR} 2016, Las Vegas, NV, USA, June 27-30, 2016}, pages
  770--778, 2016.

\bibitem{he2017}
Yihui He, Xiangyu Zhang, and Jian Sun.
\newblock Channel pruning for accelerating very deep neural networks.
\newblock In {\em {IEEE} International Conference on Computer Vision, {ICCV}
  2017, Venice, Italy, October 22-29, 2017}, pages 1398--1406, 2017.

\bibitem{mobilenetv1}
Andrew~G. Howard, Menglong Zhu, Bo Chen, Dmitry Kalenichenko, Weijun Wang,
  Tobias Weyand, Marco Andreetto, and Hartwig Adam.
\newblock Mobilenets: Efficient convolutional neural networks for mobile vision
  applications.
\newblock {\em arXiv preprint arXiv:1704.04861}, 2017.

\bibitem{hubara2016}
Itay Hubara, Matthieu Courbariaux, Daniel Soudry, Ran El{-}Yaniv, and Yoshua
  Bengio.
\newblock Quantized neural networks: Training neural networks with low
  precision weights and activations.
\newblock {\em Journal of Machine Learning Research}, 18:187:1--187:30, 2017.

\bibitem{jacob2018cvpr}
Benoit Jacob, Skirmantas Kligys, Bo Chen, Menglong Zhu, Matthew Tang, Andrew
  Howard, Hartwig Adam, and Dmitry Kalenichenko.
\newblock Quantization and training of neural networks for efficient
  integer-arithmetic-only inference.
\newblock In {\em The IEEE Conference on Computer Vision and Pattern
  Recognition (CVPR)}, June 2018.

\bibitem{koster2017flex}
Urs K{\"{o}}ster, Tristan Webb, Xin Wang, Marcel Nassar, Arjun~K. Bansal,
  William Constable, Oguz Elibol, Stewart Hall, Luke Hornof, Amir Khosrowshahi,
  Carey Kloss, Ruby~J. Pai, and Naveen Rao.
\newblock Flexpoint: An adaptive numerical format for efficient training of
  deep neural networks.
\newblock In {\em Advances in Neural Information Processing Systems 30: Annual
  Conference on Neural Information Processing Systems 2017, 4-9 December 2017,
  Long Beach, CA, {USA}}, pages 1740--1750, 2017.

\bibitem{krishnamoorthi}
Raghuraman {Krishnamoorthi}.
\newblock {Quantizing deep convolutional networks for efficient inference: A
  whitepaper}.
\newblock {\em arXiv preprint arXiv:1806.08342}, Jun 2018.

\bibitem{Li2016}
Fengfu Li and Bin Liu.
\newblock Ternary weight networks.
\newblock {\em arXiv preprint arxiv:1605.04711}, 2016.

\bibitem{liu2016ssd}
Wei Liu, Dragomir Anguelov, Dumitru Erhan, Christian Szegedy, Scott~E. Reed,
  Cheng{-}Yang Fu, and Alexander~C. Berg.
\newblock {SSD:} single shot multibox detector.
\newblock In {\em Computer Vision - {ECCV} 2016 - 14th European Conference,
  Amsterdam, The Netherlands, October 11-14, 2016, Proceedings, Part {I}},
  pages 21--37, 2016.

\bibitem{louizos2018relaxed}
Christos Louizos, Matthias Reisser, Tijmen Blankevoort, Efstratios Gavves, and
  Max Welling.
\newblock Relaxed quantization for discretized neural networks.
\newblock In {\em International Conference on Learning Representations (ICLR)},
  2019.

\bibitem{samesame}
Eldad Meller, Alexander Finkelstein, Uri Almog, and Mark Grobman.
\newblock Same, same but different: Recovering neural network quantization
  error through weight factorization.
\newblock In {\em Proceedings of the 36th International Conference on Machine
  Learning, {ICML} 2019, 9-15 June 2019, Long Beach, California, {USA}}, pages
  4486--4495, 2019.

\bibitem{mishra2017}
Asit~K. Mishra, Jeffrey~J. Cook, Eriko Nurvitadhi, and Debbie Marr.
\newblock {WRPN:} training and inference using wide reduced-precision networks.
\newblock {\em arXiv preprint arxiv 1704.03079}, 2017.

\bibitem{miyashita2016convlog}
Daisuke Miyashita, Edward~H. Lee, and Boris Murmann.
\newblock Convolutional neural networks using logarithmic data representation.
\newblock {\em arXiv preprint arxiv:1603.01025}, 2016.

\bibitem{nair2010}
Vinod Nair and Geoffrey~E. Hinton.
\newblock Rectified linear units improve restricted boltzmann machines.
\newblock In {\em Proceedings of the 27th International Conference on Machine
  Learning (ICML-10), June 21-24, 2010, Haifa, Israel}, pages 807--814, 2010.

\bibitem{pytorch}
Adam Paszke, Sam Gross, Soumith Chintala, Gregory Chanan, Edward Yang, Zachary
  DeVito, Zeming Lin, Alban Desmaison, Luca Antiga, and Adam Lerer.
\newblock Automatic differentiation in pytorch.
\newblock 2017.

\bibitem{peters2018}
Jorn W.~T. Peters and Max Welling.
\newblock Probabilistic binary neural networks.
\newblock {\em arXiv preprint arxiv:1809.03368}, 2018.

\bibitem{rastegari2016}
Mohammad Rastegari, Vicente Ordonez, Joseph Redmon, and Ali Farhadi.
\newblock Xnor-net: Imagenet classification using binary convolutional neural
  networks.
\newblock In {\em Computer Vision - {ECCV} 2016 - 14th European Conference,
  Amsterdam, The Netherlands, October 11-14, 2016, Proceedings, Part {IV}},
  pages 525--542, 2016.

\bibitem{ILSVRC15}
Olga Russakovsky, Jia Deng, Hao Su, Jonathan Krause, Sanjeev Satheesh, Sean Ma,
  Zhiheng Huang, Andrej Karpathy, Aditya Khosla, Michael Bernstein,
  Alexander~C. Berg, and Li Fei-Fei.
\newblock {ImageNet Large Scale Visual Recognition Challenge}.
\newblock {\em International Journal of Computer Vision (IJCV)},
  115(3):211--252, 2015.

\bibitem{mobilenetv2}
Mark Sandler, Andrew Howard, Menglong Zhu, Andrey Zhmoginov, and Liang-Chieh
  Chen.
\newblock Mobilenetv2: Inverted residuals and linear bottlenecks.
\newblock In {\em The IEEE Conference on Computer Vision and Pattern
  Recognition (CVPR)}, June 2018.

\bibitem{sheng2018}
Tao Sheng, Chen Feng, Shaojie Zhuo, Xiaopeng Zhang, Liang Shen, and Mickey
  Aleksic.
\newblock A quantization-friendly separable convolution for mobilenets.
\newblock In {\em 1st Workshop on Energy Efficient Machine Learning and
  Cognitive Computing for Embedded Applications (EMC2)}, 2018.

\bibitem{stock2018equinormalization}
Pierre Stock, Benjamin Graham, Rémi Gribonval, and Hervé Jégou.
\newblock Equi-normalization of neural networks.
\newblock In {\em International Conference on Learning Representations}, 2019.

\bibitem{ullrich2018}
Karen Ullrich, Edward Meeds, and Max Welling.
\newblock Soft weight-sharing for neural network compression.
\newblock In {\em International Conference on Learning Representations (ICLR)},
  2017.

\bibitem{zhang2015}
Xiangyu Zhang, Jianhua Zou, Kaiming He, and Jian Sun.
\newblock Accelerating very deep convolutional networks for classification and
  detection.
\newblock {\em {IEEE} Trans. Pattern Anal. Mach. Intell.}, 38(10):1943--1955,
  2016.

\bibitem{Zhou2017}
Aojun Zhou, Anbang Yao, Yiwen Guo, Lin Xu, and Yurong Chen.
\newblock Incremental network quantization: Towards lossless cnns with
  low-precision weights.
\newblock {\em arXiv preprint arxiv:1702.03044}, abs/1702.03044, 2017.

\bibitem{Zhou2016}
Shuchang Zhou, Zekun Ni, Xinyu Zhou, He Wen, Yuxin Wu, and Yuheng Zou.
\newblock Dorefa-net: Training low bitwidth convolutional neural networks with
  low bitwidth gradients.
\newblock {\em arXiv preprint arXiv:1606.06160}, 2016.

\end{thebibliography}
